\documentclass[conference]{IEEEtran}
\IEEEoverridecommandlockouts

\usepackage{cite}
\usepackage{amsmath,amssymb,amsfonts}
\usepackage{algorithm}
\usepackage{algpseudocode}
\usepackage{graphicx}
\usepackage{textcomp}
\usepackage[table]{xcolor}
\usepackage[colorlinks=true,urlcolor=blue]{hyperref}
\usepackage{booktabs}
\usepackage{multirow}
\usepackage{threeparttable}

\newcommand{\omicron}{o}

\def\BibTeX{{\rm B\kern-.05em{\sc i\kern-.025em b}\kern-.08em
    T\kern-.1667em\lower.7ex\hbox{E}\kern-.125emX}}
\begin{document}

\title{CITRAS-FM: Tiny Time Series Foundation Model for Covariate-Informed Zero-Shot Forecasting\thanks{Accepted to EUSIPCO 2026.}}

\author{
  \IEEEauthorblockN{
    Yosuke Yamaguchi,
    Issei Suemitsu,
    Yuki Kajihara,
    Wenpeng Wei
  }
  \IEEEauthorblockA{
    Research \& Development Group, Hitachi Ltd., Tokyo, Japan\\
    \{yosuke.yamaguchi.fy, issei.suemitsu.rj, yuki.kajihara.fj, wenpeng.wei.bo\}@hitachi.com
  }
}

\maketitle

\begin{abstract}
Pretrained time series foundation models (TSFMs) have enabled zero-shot forecasting on unseen target series. However, existing TSFMs often incur high computational cost and provide limited support for diverse variable types, often failing to account for covariates that exogenously influence target variability. To address these challenges, we propose CITRAS‑FM, a tiny 7M-parameter TSFM that supports univariate, multivariate, and covariate‑informed zero-shot forecasting with real-time CPU inference. Built on a patch‑based, decoder‑only Transformer, CITRAS‑FM introduces Shifted Attention into the cross‑variate module to effectively exploit known covariates accessible throughout the forecast horizon. Moreover, to enable covariate‑aware pretraining despite the scarcity of covariate-rich corpora, we propose CovSynth, which synthesizes realistic covariates from decomposed components of target series. Experiments on fev‑bench, spanning 100 tasks across various settings, demonstrate that CITRAS‑FM achieves state-of-the-art zero-shot accuracy among sub‑10M TSFMs while delivering sub‑0.1‑second CPU inference, offering a strong balance between forecasting accuracy and real-time deployability. The code is available at \url{https://github.com/hitachi-ais/citras-fm}.
\end{abstract}

\bstctlcite{IEEEexample:BSTcontrol}
\section{Introduction}

Time series forecasting---predicting the future values of target variables---is widely used across domains such as server load forecasting~\cite{cohen2025toto} and energy demand forecasting~\cite{thamer2025energydemand}. Recently, time series foundation models~(TSFMs)~\cite{das2023timesfm} have gained traction. These models are trained on massive datasets to learn generalizable temporal dynamics, enabling zero-shot forecasting on unseen datasets without task-specific training. This paradigm is particularly promising in industrial environments where data distributions shift continually and collecting sufficient historical data for each deployment is impractical. However, two shortcomings limit the practical use of existing TSFMs.

First, most leading TSFMs on popular benchmarks are computationally expensive. For example, TimesFM-2.5~\cite{das2023timesfm} and Chronos-2~\cite{ansari2025chronos2} contain over a hundred million parameters, resulting in slow inference on limited computational resources. Many real‑world systems—server clusters or sensor‑dense manufacturing equipment—require on‑device low‑latency operation~\cite{sandur2022jarvis}. Although lightweight TSFMs exist, they typically lag behind larger models in forecasting accuracy~\cite{ansari2024chronos} and often require fine‑tuning to achieve competitive performance~\cite{ekambaram2024ttm}.

Second, the majority of existing TSFMs are restricted to relatively simple univariate settings.
In practice, applications frequently demand simultaneous forecasting of multiple targets (\emph{multivariate}), or the incorporation of covariates that represent exogenous signals.
For instance, in server metric forecasting, it is crucial to maintain consistency between CPU temperature and power consumption~\cite{cohen2025toto}; in energy demand forecasting, considering the impact of temperature and holiday calendar is essential~\cite{yamaguchi2025citras}. 
However, TSFM architectures that can accommodate such diverse variables remain largely underexplored. Furthermore, large‑scale pretraining corpora containing rich covariates are scarce~\cite{aksu2024gifteval}.
Consequently, as summarized in Table~\ref{tab:tsfms}, most TSFMs support only a limited subset of variable types in the zero-shot setting. To the best of our knowledge, Chronos-2~\cite{ansari2025chronos2} is the only model that supports zero‑shot forecasting with all of these variable types, but its 120M parameters impose heavy computational demands, limiting its practicality in resource-constrained environments.

To address these limitations, we propose \textbf{CITRAS‑FM}, a \textbf{C}ovariate‑\textbf{I}nformed \textbf{Tra}n\textbf{s}former for time series \textbf{F}oundation \textbf{M}odeling. CITRAS‑FM is a \emph{tiny} 7M‑parameter TSFM that enables zero‑shot forecasting in univariate, multivariate, and covariate-informed settings, and provides fast CPU inference suitable for real-time application. It builds on our prior work, CITRAS~\cite{yamaguchi2025citras}, which enables flexible handling of different variable types under supervised settings.
To enhance covariate-informed zero‑shot forecasting performance, CITRAS‑FM introduces a novel \emph{Shifted Attention} layer in the cross‑variate module of CITRAS, enabling effective use of covariates available throughout the forecast horizon for target prediction. Furthermore, to address the scarcity of covariates in existing pretraining corpora, we propose \emph{CovSynth}, a method that synthesizes pseudo‑covariates that capture target fluctuations using decomposed components of the target series. These innovations enable pretraining CITRAS‑FM for zero‑shot covariate‑informed forecasting using large‑scale datasets that predominantly contain target‑only time series.

Experiments on \textit{fev‑bench}~\cite{shchur2025fevbench}, covering 100 tasks that span univariate, multivariate, and covariate‑informed settings, show that CITRAS‑FM achieves the best zero‑shot forecasting accuracy among sub‑10\,M tiny TSFMs. Moreover, it outperforms TSFMs with over $20\times$ more parameters (e.g., TimesFM‑2.5~\cite{das2023timesfm}, COSMIC~\cite{auer2025cosmic}) under covariate‑informed settings. In terms of computational efficiency, it achieves sub‑0.1‑second CPU inference in covariate‑informed scenarios, offering a favorable balance between flexible variable utilization and real-time deployability.

\begin{table}[t]
\centering
\begin{threeparttable}
\caption{Supported Variable Types for Zero-shot Forecasting}
\label{tab:tsfms}
\begin{tabular}{l c c c c}
\toprule
\multirow{2}{*}{Models} & \multirow{2}{*}{Univariate} & \multirow{2}{*}{Multivariate} &
\multicolumn{1}{c}{Observed\tnote{1}} &
\multicolumn{1}{c}{Known\tnote{2}} \\
 &  &  & \multicolumn{1}{c}{covariate} & \multicolumn{1}{c}{covariate} \\
\midrule
CITRAS-FM (ours)     & \checkmark & \checkmark & \checkmark & \checkmark \\
Chronos-2~\cite{ansari2025chronos2}     & \checkmark & \checkmark & \checkmark & \checkmark \\
Toto-1.0~\cite{cohen2025toto}      & \checkmark & \checkmark & \checkmark & \(\times\) \\
TabPFN-TS~\cite{hoo2025tabpfn}     & \checkmark & \(\times\)  & \(\times\)  & \checkmark \\
COSMIC~\cite{auer2025cosmic}        & \checkmark & \(\times\)  & \checkmark & \checkmark \\
\midrule
TimesFM-2.5~\cite{das2023timesfm}   & \checkmark & \(\times\)  & \(\times\)  & \(\times\) \\
Sundial~\cite{liu2025sundial}       & \checkmark & \(\times\)  & \(\times\)  & \(\times\) \\
Moirai-2.0~\cite{liu2025moirai2}    & \checkmark & \(\times\)  & \(\times\)  & \(\times\) \\
Chronos-Bolt~\cite{ansari2024chronos}  & \checkmark & \(\times\)  & \(\times\)  & \(\times\) \\
YINGLONG~\cite{wang2025yinglong}      & \checkmark & \(\times\)  & \(\times\)  & \(\times\) \\
KAIROS~\cite{feng2025kairos}        & \checkmark & \(\times\)  & \(\times\)  & \(\times\) \\
TinyTimeMixer~\cite{ekambaram2024ttm} & \checkmark & \(\times\)  & \(\times\)  & \(\times\) \\
\bottomrule
\end{tabular}
\begin{tablenotes}[flushleft]\footnotesize
\item[1] \emph{Observed covariate} has recorded values up to the prediction point.
\item[2] \emph{Known covariate} has values from the past through to the forecast horizon.
\end{tablenotes}
\end{threeparttable}
\end{table}

\section{Methodology}
\subsection{Problem Setting}
Let $\mathbf{X}_{1:T}^{\tau,:}=\{ \mathbf{X}_{1:T}^{\tau,1},\mathbf{X}_{1:T}^{\tau,2}, \ldots, \mathbf{X}_{1:T}^{\tau,C_{\tau}} \} \in \mathbb{R}^{T \times C_{\tau}}$ be a target time series of length $T$ and $C_{\tau}$ variables. 
We consider settings where \emph{observed covariates} $\mathbf{X}_{1:T}^{\omicron,:} \in \mathbb{R}^{T \times C_{\omicron}}$, which are available up to the prediction point, and \emph{known covariates} $\mathbf{X}_{1:T+S}^{\kappa,:} \in \mathbb{R}^{(T+S) \times C_{\kappa}}$, which are available throughout the forecast horizon, may be present.
Known covariates can take predetermined (e.g., calendar events), planned (e.g., promotional schedules), or forecasted (e.g., weather forecasts) values. 
The objective is to estimate the conditional distribution of the future target values $P\left( \mathbf{X}_{T+1:T+S}^{\tau,:} \mid \mathbf{X}_{1:T}^{\tau,:},\ \mathbf{X}_{1:T}^{\omicron,:},\ \mathbf{X}_{1:T+S}^{\kappa,:} \right)$.

\subsection{Architecture}
\begin{figure}[t]
    \centering
    \includegraphics[trim=0 2 0 0, clip, width=0.85\columnwidth]{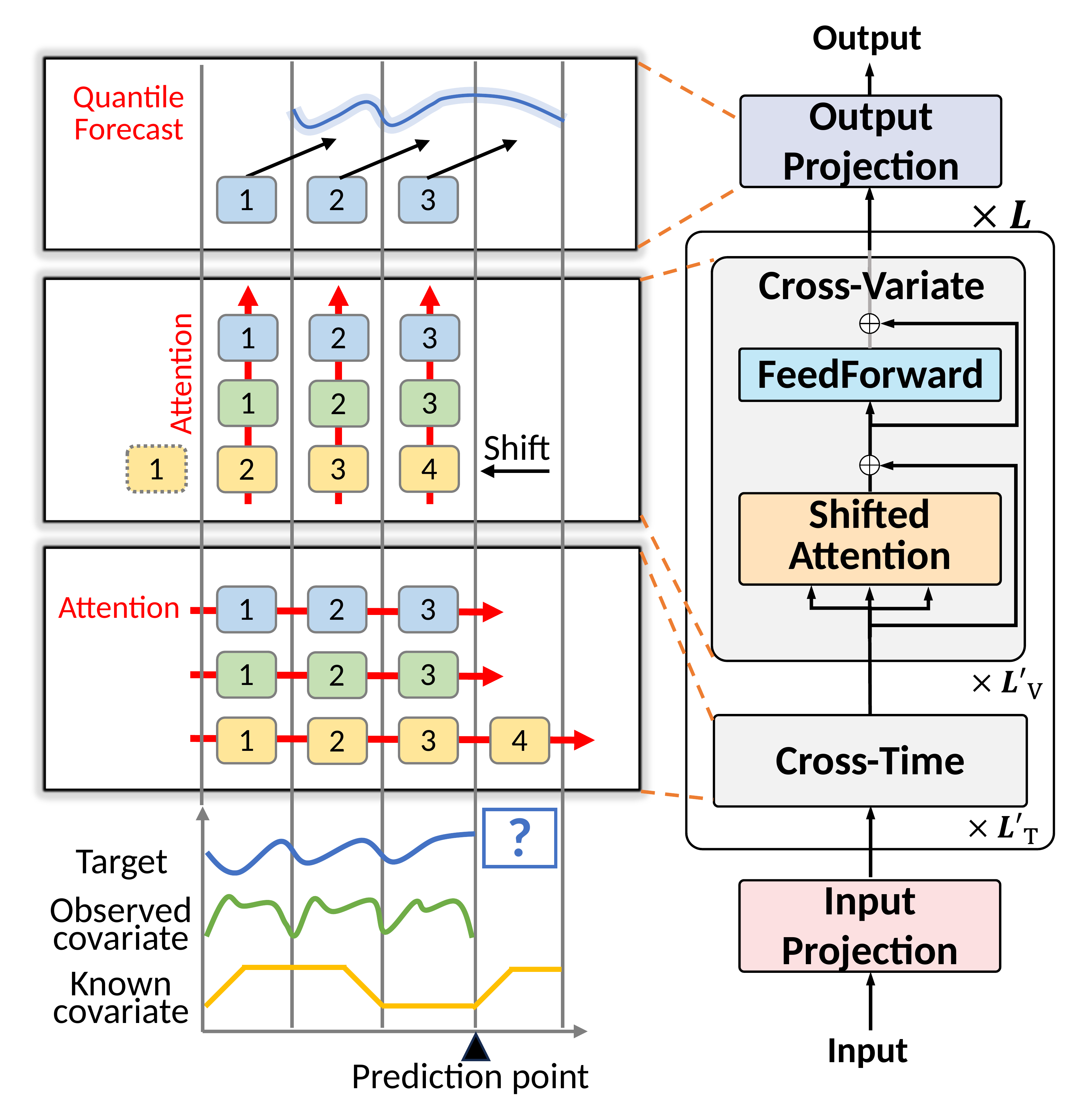}
    \caption{Overall structure of CITRAS-FM.}
    \label{fig:arc}
\end{figure}

The overall architecture of CITRAS-FM is shown in Fig.~\ref{fig:arc}. The architecture follows our prior work, CITRAS~\cite{yamaguchi2025citras}, a patch-based decoder-only Transformer~\cite{vaswani2017attention} originally designed for supervised forecasting, and consists of four modules: input projection, cross-time attention, cross-variate attention, and output projection. CITRAS is designed to model diverse variables while preventing any unintended leakage of future information at any patch step, thereby allowing training with dense next‑token prediction signals.
In this work, we introduce a new Shifted Attention layer in the cross‑variate attention module to improve generalization in capturing covariate information.
Moreover, we incorporate recent techniques such as causal scaling~\cite{das2023timesfm}, pre-layer normalization~\cite{xiong2020prelayernorm}, and SwiGLU activation~\cite{shazeer2020swiglu} for improved stability, and a quantile regression head for probabilistic forecasting.

\textbf{Input Projection.} This module converts raw time series into token embeddings. To preserve the local semantics of time series, we first apply patching~\cite{nie2022patchtst}, where each variable is segmented into non-overlapping patches of length $P$. When $T$ and $S$ are not divisible by $P$, we apply zero-padding on the left and right, respectively.
Then we apply patch-wise causal scaling~\cite{das2023timesfm} to mitigate non-stationarity while preventing information leakage from future patches. 
Taking a target variable $c$ as an example, this can be formalized as:
\begin{equation}
\begin{aligned}
    \{\mathbf{s}_1^{\tau,c},\mathbf{s}_2^{\tau,c}, \ldots,\mathbf{s}_{N_{\tau}}^{\tau,c} \} = \operatorname{CausalScale} \left(\operatorname{Patchify} \left(\mathbf{X}_{1:T}^{\tau,c} \right) \right)\\
\end{aligned}    
\end{equation}
where $N_{\tau} = \frac{T}{P}$ is the number of patches.
Then each patch $\mathbf{s}_i^{\tau,c} \in \mathbb{R}^P$ is concatenated with a binary padding mask $\mathbf{m}_i^{\tau,c}$ (1 for padded positions), and projected to a token embedding:
\begin{equation}
    \mathbf{H}_i^{\tau,c} = \operatorname{Embed}\left(\left[\mathbf{s}_i^{\tau,c}, \mathbf{m}_i^{\tau,c}\right]\right), ~~i=1, \ldots,N_{\tau}
\end{equation}
where $\operatorname{Embed}: \mathbb{R}^{2P} \to \mathbb{R}^D$ is a residual network with one hidden layer.
We denote the token embeddings of a target variable at all patch steps as $ \mathbf{H}_:^{\tau,c}=~\{ \mathbf{H}_i^{\tau,c} \}_{i=1}^{N_{\tau}}~\in~\mathbb{R}^{N_{\tau} \times D}$. Similarly, token embeddings for observed covariates $\mathbf{H}_:^{\omicron,c} \in \mathbb{R}^{N_{\omicron} \times D}$, and known covariates $\mathbf{H}_:^{\kappa,c} \in \mathbb{R}^{N_{\kappa} \times D}$ are obtained using shared parameters with the target variables. Here, $N_{\omicron} = N_{\tau} = \frac{T}{P}$ and $N_{\kappa} =\frac{(T+S)}{P}$.

\textbf{Cross-Time Attention.} This module captures cross-time dependencies of each variable separately via multi-head attention with causal masking.
We adopt Rotary Position Embedding (RoPE) \cite{jianlin2024roformer} to encode position information and apply pre-layer normalization \cite{xiong2020prelayernorm} for improved training stability.  
Taking a target variable $c$ as an example, this can be formalized as:
\begin{equation}
\begin{aligned}
\widetilde{\mathbf{H}}_:^{\tau,c}
&= \mathbf{H}_:^{\tau,c}
   + \operatorname{MHA}\!\big(
       \operatorname{LN}(\mathbf{H}_:^{\tau,c}),\,
       \operatorname{LN}(\mathbf{H}_:^{\tau,c}),\,
       \operatorname{LN}(\mathbf{H}_:^{\tau,c})
     \big),\\
\mathbf{H}_:^{\tau,c}
&= \widetilde{\mathbf{H}}_:^{\tau,c}
   + \operatorname{FFN}\!\big(\operatorname{LN}(\widetilde{\mathbf{H}}_:^{\tau,c})\big),
\end{aligned}
\end{equation}
where $\operatorname{LN}$ denotes layer normalization, $\operatorname{MHA}(\mathbf{Q},\mathbf{K},\mathbf{V})$ denotes the multi-head attention layer with queries $\mathbf{Q}$, keys $\mathbf{K}$, and values $\mathbf{V}$, and $\operatorname{FFN}$ denotes a feed-forward network with a SwiGLU‑based gated activation \cite{shazeer2020swiglu}.
This way, each token embedding at patch step $i$ summarizes information from all preceding patches of the same variable.
The same operations are applied to all observed and known covariates separately.

\textbf{Cross-Variate Attention.}
This module captures information from other variables that is useful for next‑token (patch) prediction of the target variables.
At this stage, it is important to incorporate not only the target and observed covariates at the same patch step, but also the known covariates from one patch ahead, 
as they carry forecasting‑horizon information that may directly influence the target values.
Consequently, conventional patch‑wise cross‑variate attention restricted to the same patch step becomes insufficient.
CITRAS~\cite{yamaguchi2025citras} overcomes this problem by KV Shift, which associates the attention key of the known covariate with the value from one patch step ahead.
Although effective in supervised settings, a simpler form of temporal alignment is desirable for more generalized utilization of covariate information.
To this end, we introduce a novel Shifted Attention layer in CITRAS-FM.
In this layer, the first token embedding of the known covariates is removed, and the remaining token embeddings are shifted left by one patch. As a result, at each patch step, the target token can directly attend not only to the current-step target and observed covariate tokens, but also to the next-step known covariate token.
Since the next-step token already contains information from all preceding patches, the first token can be safely removed without significant loss of information.

At patch step $i$, this operation can be formalized as:
\begin{equation}
\begin{aligned}
\widetilde{\mathbf{H}}_i^{\tau,:}
&= \mathbf{H}_i^{\tau,:}
   + \operatorname{MHA}\!\big(
       \operatorname{LN}(\mathbf{H}_i^{\tau,:}),\,
       \operatorname{LN}(\mathbf{H}_i^{kv,:}),\,
       \operatorname{LN}(\mathbf{H}_i^{kv,:})
     \big),\\
\mathbf{H}_i^{\tau,:}
&= \widetilde{\mathbf{H}}_i^{\tau,:}
   + \operatorname{FFN}\!\big(\operatorname{LN}(\widetilde{\mathbf{H}}_i^{\tau,:})\big),
\end{aligned}
\end{equation}
where $i=1, \ldots,N_{\tau}$ and 
\begin{equation}
\begin{aligned}
    \mathbf{H}_i^{kv,:} &= \left[ \mathbf{H}_i^{\tau,:},\, \mathbf{H}_i^{\omicron,:},\, \mathbf{H}_{i+1}^{\kappa,:} \right]  \in \mathbb{R}^{(C_{\tau}+C_{\omicron}+C_{\kappa}) \times D}\\
\end{aligned}
\end{equation}

\textbf{Output Projection.} 
In this module, each target token embedding is used to generate the forecast of the next patch. 
To facilitate probabilistic forecasting, we output 9 quantiles $Q=\{0.1, 0.2, \ldots, 0.9\}$ for each time step in the patch.
For $i=1, \ldots,N_{\tau}$ and $c=1, \ldots,C_{\tau}$, this can be formalized as:
\begin{equation}
     \widehat{\mathbf{X}}_{iP+1:(i+1)P}^{\tau,c} = \operatorname{CausalRescale}\left( \operatorname{Project}\left( \mathbf{H}_i^{\tau,c} \right) \right)
\end{equation}
where $\widehat{\mathbf{X}}_{iP+1:(i+1)P}^{\tau,c} \in \mathbb{R}^{P \times |Q|}$, $\operatorname{Project}: \mathbb{R}^D \to \mathbb{R}^{P \times |Q|}$ is a shared residual network as used in~\cite{ansari2025chronos2,liu2025moirai2}, and $\operatorname{CausalRescale}$ is the inverse operation of $\operatorname{CausalScale}$.

The model is trained to minimize the quantile loss~\cite{ansari2025chronos2} using all of the predicted patches, following the next-token prediction scheme.
In the testing phase, the model can autoregressively generate forecasts for horizon $S$ by recursively feeding the predicted patches into subsequent inputs when $S>P$.
 
\subsection{Pretraining}

We use three time series datasets for pretraining: \emph{TSMixup}~\cite{ansari2024chronos}, the \emph{Cauker}-generated dataset~\cite{xie2025cauker}, and the \emph{Gift-Eval pretraining} dataset~\cite{aksu2024gifteval}. TSMixup is an augmented univariate corpus used in Chronos~\cite{ansari2024chronos} and contains 11B time points. Cauker is a time-series synthesis method that reproduces dependencies among multiple series using structural causal models (SCMs). We generate 4B time points with up to 15 variables using Cauker, where each variable was randomly assigned as a target, an observed covariate, or a known covariate at each training step. For the Gift-Eval pretraining dataset, we adopt the same subset as in Chronos-2~\cite{ansari2025chronos2}, which contains 19B time points of univariate target series.

However, these datasets have limited quantity and diversity of covariates. To address this, we introduce CovSynth, a procedure that synthesizes plausible covariates directly from target series, enabling covariate-aware pretraining without dedicated covariate-rich datasets. The core idea behind CovSynth is that effective covariates should contribute information that is not well explained by the target's own temporal dynamics (e.g., trend and seasonality). Accordingly, CovSynth first decomposes the target into trend, seasonal, and residual components via STL decomposition~\cite{cleveland1990stl}. It then utilizes the residual component---i.e., variability not captured by trend or seasonality---to generate three types of covariates that mimic common real-world exogenous drivers.

\textbf{Event covariate.} This mimics short, irregular events that abruptly affect the target (e.g., large festivals affecting regional electricity demand, or batch jobs that spike server load).
It is generated by applying a quantile-based threshold to the residual component and binarizing the series into 0/1 values.

\textbf{Long-term covariate.} This mimics exogenous signals underlying long-horizon variations correlated with the target (e.g., macroeconomic indicators influencing consumer spending, or climate oscillation indices affecting agricultural production).
It is generated by taking a weighted combination of the residual and trend components.

\textbf{Periodic covariate.} This mimics exogenous signals underlying periodic variations correlated with the target (e.g., hourly electricity demand affecting prices, solar irradiance affecting PV output).
It is generated by taking a weighted combination of the residual and seasonal components.

\begin{figure}[t]
    \centering
    \includegraphics[trim=8 0 0 0, clip, width=1.0\columnwidth]{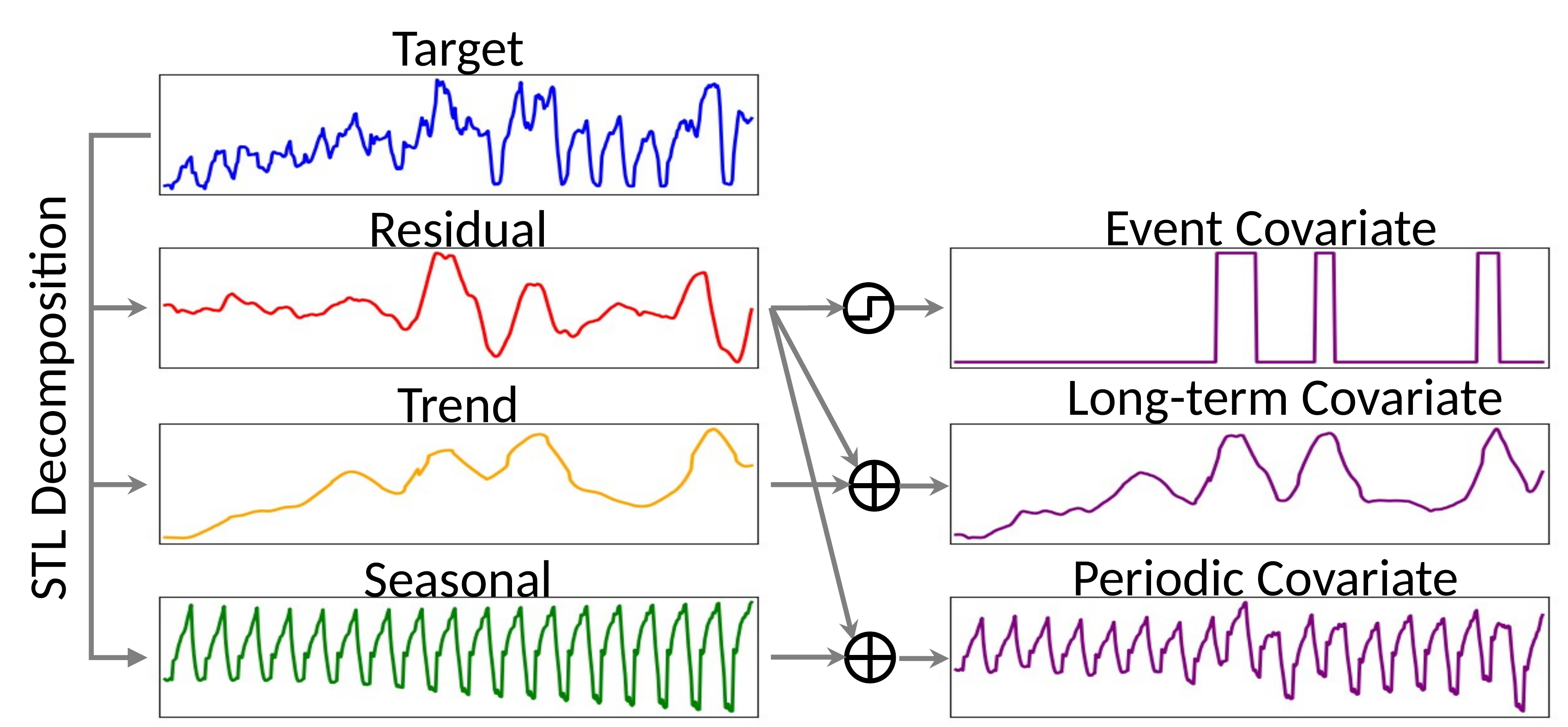}
    \caption{Procedure of CovSynth.}
    \label{fig:covsynth}
\end{figure}

Fig.~\ref{fig:covsynth} illustrates an example of synthesizing covariates from an actual target series in the Gift-Eval pretraining dataset using CovSynth. We observe that the residual captures variability in the target that is difficult to predict from its own temporal patterns, and the three synthesized covariates emulate external signals that reflect this variability in different ways.

We apply CovSynth to the target series from the Gift-Eval pretraining dataset. Specifically, for each sampled target series at every training step, we synthesize up to five observed covariates and five known covariates by randomly varying the quantile threshold, mixing weights, and additional noise levels. Through this process, CITRAS-FM is pretrained on datasets enriched with diverse covariate patterns.

\subsection{Hyperparameters}

For the architecture, we set the patch size to $P=24$, the model dimension to $D=256$, and the number of attention heads to $8$. 
The cross-time and cross-variate attention modules are stacked with $L'_T=3$ and $L'_V = 1$, and this block is repeated $L=2$ times. 
For pretraining, we use a maximum input length of $1032$, a batch size of $256$, and train for $500{,}000$ steps.
Optimization is performed using the AdamW optimizer (learning rate $1\!\times\!10^{-4}$, weight decay $0.01$, $\beta_1=0.9$, $\beta_2=0.999$).
The learning rate schedule consists of a linear warmup over the first $10{,}000$ steps followed by cosine annealing.
For each batch instance, the target series is sampled from TSMixup, the Cauker-generated dataset, and the Gift-Eval pretraining dataset with equal probability.
A single NVIDIA V100 (32\,GB) GPU is used for pretraining.

\section{Experiments}
\subsection{Benchmark Results}
We first evaluate the zero-shot forecasting accuracy of CITRAS-FM on fev-bench~\cite{shchur2025fevbench}. This benchmark comprises 100 forecasting tasks spanning diverse domains, seasonalities, and forecast horizons, enabling a comprehensive evaluation under realistic conditions. We categorize the tasks into four groups: \emph{fev-all} (all 100 tasks), \emph{fev-cov} (42 tasks with observed and/or known covariates), \emph{fev-multi} (26 tasks with multivariate targets), and \emph{fev-uni} (32 tasks with univariate targets).
Following prior work~\cite{ansari2025chronos2}, we compute Scaled Quantile Loss (SQL) for each task to evaluate probabilistic forecasting performance and aggregate these scores across tasks by \emph{skill score} (average percentage improvement over a SeasonalNaive baseline).

Table~\ref{tab:fev} presents the skill scores for each group. Among tiny TSFMs under 10M parameters, CITRAS-FM achieves the highest skill scores across all groups, with a margin of 3.5 points over the second-best $\text{KAIROS}_{\text{mini}}$ on fev-all. Even when compared with larger models, CITRAS-FM attains top-tier accuracy on fev-cov, outperforming TSFMs with more than 20$\times$ larger parameter counts (e.g., TimesFM-2.5, COSMIC). These results highlight the effectiveness of CITRAS-FM in leveraging covariate information for forecasting. Although Chronos-2 and TabPFN-TS achieve higher scores in fev-cov, these two models incur substantially higher inference latency, as shown in the next section. This makes them less suitable for real‑time deployment in resource‑constrained settings, which is the focus of this work.

\begin{figure}[h]
    \centering
    \includegraphics[trim=0 0 0 8, clip, width=1.0\columnwidth]{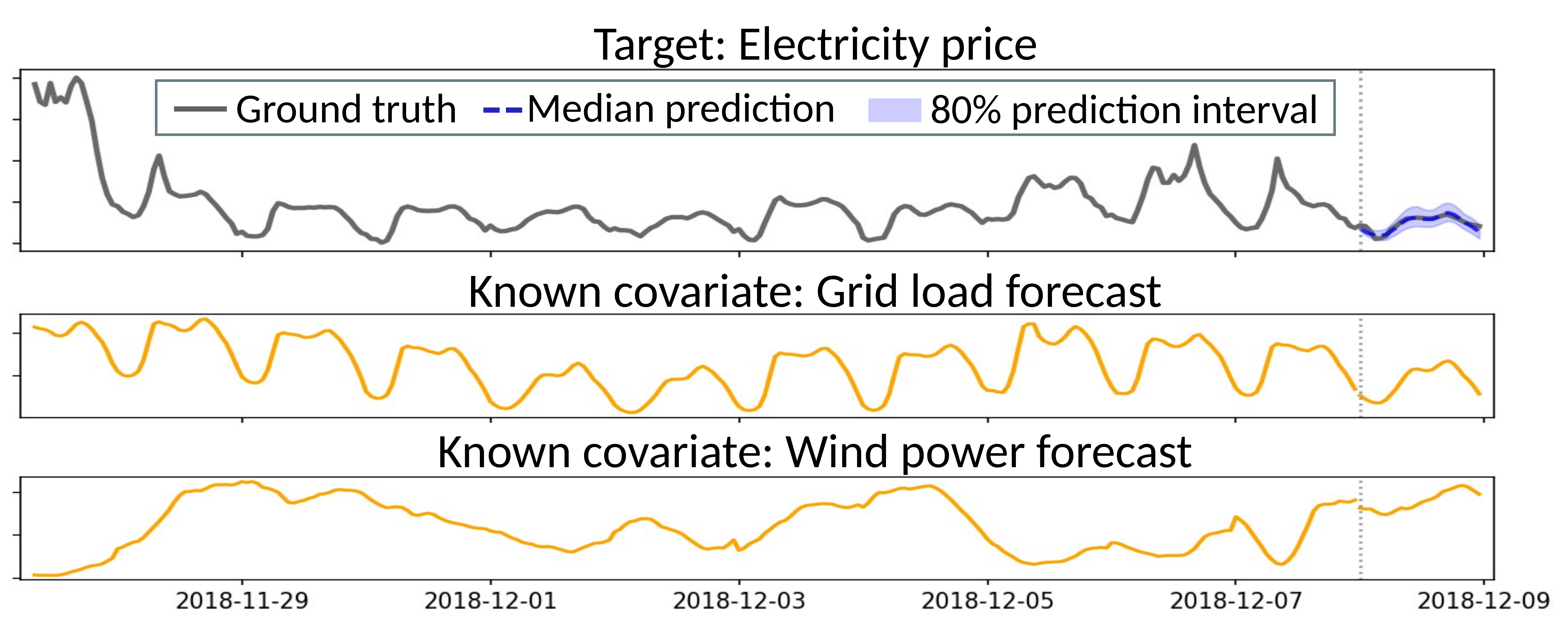}
    \caption{Zero-shot forecasting results on EPF-NP.}
    \label{fig:epf_np}
\end{figure}

\textbf{Qualitative Results.}
Fig.~\ref{fig:epf_np} illustrates forecasting results of CITRAS-FM on EPF-NP, a day-ahead hourly electricity price forecasting task in fev-bench. Along with target prices, this task provides public grid-load and wind-power forecasts as known covariates. CITRAS-FM accurately anticipates the price decline over the forecast horizon, considering decreasing grid load and increasing wind generation. Importantly, this market-driven dynamic is captured in a zero-shot manner from historical input, without task-specific finetuning.

\begin{table}
\caption{Results on fev-bench within each parameter range. The best results are highlighted in bold, followed by underline.}
\label{tab:fev}
\centering
\begin{tabular}{l r r r r r}
\toprule
 & Params. & \multicolumn{4}{c}{Skill Score (\%) $\uparrow$ }\\
Model & \multicolumn{1}{c}{(M)} & fev-all & fev-cov & fev-multi & fev-uni \\
\midrule
\multicolumn{6}{l}{\textit{Models with $<$10M parameters}} \\
\midrule
\textit{CITRAS-FM} & 7.2 & \textbf{41.2} & \textbf{39.0} & \textbf{54.2} & \textbf{31.3} \\
$\text{KAIROS}_{\text{mini}}$ & 9.9 & \underline{37.7} & \underline{35.4} & \underline{50.9} & \underline{27.9} \\
$\text{Chronos-Bolt}_{\text{Tiny}}$ & 8.7 & 35.9 & 32.9 & 49.8 & 26.4 \\
$\text{YINGLONG}_{6m}$ & 7.3 & 25.1 & 26.5 & 49.7 & -6.2 \\
$\text{TinyTimeMixer}_{\text{r2}}$ & 0.8 & -1.1 & -4.6 & 28.2 & -27.5 \\
SeasonalNaive & 0 & 0.0 & 0.0 & 0.0 & 0.0 \\
\midrule
\multicolumn{6}{l}{\textit{Models with $\geq$10M parameters}} \\
\midrule
Chronos-2 & 120 & \textbf{47.3} & \textbf{47.0} & \textbf{57.9} & \textbf{37.0} \\
TiRex & 35 & \underline{42.6} & 38.7 & 55.7 & \underline{35.0} \\
TimesFM-2.5 & 200 & 42.3 & 37.6 & 56.1 & 34.9 \\
Toto-1.0 & 151 & 40.7 & 35.1 & \underline{57.1} & 31.6 \\
TabPFN-TS & 11 & 39.6 & \underline{40.0} & 47.7 & 31.4 \\
$\text{Chronos-Bolt}_{\text{Base}}$ & 205 & 38.9 & 35.9 & 52.1 & 30.1 \\
COSMIC & 200 & 39.0 & 36.0 & 52.3 & 29.8 \\
Moirai-2.0 & 11 & 39.3 & 36.6 & 53.4 & 29.0 \\
$\text{Sundial}_{\text{Base}}$ & 128 & 33.4 & 28.0 & 51.1 & 22.9 \\
\bottomrule
\end{tabular}
\end{table}

\subsection{Efficiency Comparison}
In practical deployments of forecasting models, fast inference in resource-constrained environments is often required. As a representative scenario, we evaluate models on the \emph{Application} dataset~\cite{ekambaram2024ttm}, comparing forecasting accuracy and CPU inference speed. This dataset consists of monitored metrics from the “Stan’s Robot Shop’’ e-commerce application, and includes four business KPIs (e.g., payment, catalog) along with 35 IT event metrics. Following \cite{ekambaram2024ttm}, the business KPIs serve as target variables and the IT event metrics are treated as known covariates. All signals are monitored at a 10-second resolution, implying near-real-time responsiveness is desirable.

We utilize the rolling-window evaluation protocol, where the number of evaluation windows is 20, each window’s forecasting horizon is 24 steps (4 minutes), and the stride between consecutive windows is also 24 steps. To enable a fair comparison of inference speed, we fix the historical input length to 1{,}032 for all models. CPU inference time is measured on an \texttt{Intel(R) Core(TM) i5-14400F}.

Table~\ref{tab:app} summarizes the average SQL and the average CPU inference time per window (wall-clock).
We observe that CITRAS-FM, $\text{KAIROS}_{\text{mini}}$, and $\text{Chronos-Bolt}_{\text{Tiny}}$ yield sub-0.1-second inference per window, making them suitable for real-time applications. Among these three, \emph{only} CITRAS-FM leverages future-known IT event covariates and achieves a substantial improvement in SQL. In contrast, Chronos‑2 requires more than $30\times$ longer inference time than CITRAS‑FM due to its large parameter count. TabPFN‑TS requires over $900\times$ longer inference time because it formulates forecasting as tabular regression and recomputes a heavy transformer‑based posterior approximation for every prediction. Overall, CITRAS‑FM provides an effective balance between covariate utilization and real-time deployability.

\begin{table}
\caption{Efficiency Analysis}
\label{tab:app}
\centering
\begin{tabular}{l r r}
\toprule
Model & SQL (\%) $\downarrow$ & Inference Time (sec) $\downarrow$ \\
\midrule
CITRAS-FM & 1.09 & 0.05 \\
Chronos-2 & 0.76 & 1.79 \\
TabPFN-TS & 1.38 & 46.52 \\
$\text{KAIROS}_{\text{mini}}$ & 2.51 & 0.07 \\
$\text{Chronos-Bolt}_{\text{Tiny}}$ & 2.09 & 0.06 \\
\bottomrule
\end{tabular}
\end{table}

\subsection{Ablation}
Table~\ref{tab:ablation} presents the ablation results. Removing Shifted Attention prevents the model from leveraging known covariates, resulting in a 2.6-point drop in skill score on fev-cov. Excluding CovSynth reduces the diversity of covariates during pretraining, leading to a 1.8-point decrease. These findings demonstrate that both components play essential roles in strengthening CITRAS-FM's covariate modeling capability.
\begin{table}
\caption{Ablation}
\label{tab:ablation}
\centering
\begin{tabular}{l r r r r}
\toprule
 & \multicolumn{4}{c}{Skill Score (\%) $\uparrow$} \\
Model & fev-all & fev-cov & fev-multi & fev-uni \\
\midrule
CITRAS-FM (full) & \textbf{41.2} & \textbf{39.0} & \textbf{54.2} & 31.3 \\
w/o Shifted Attention & 39.9 & 36.4 & 53.7 & 31.0 \\
w/o CovSynth & 40.5 & 37.2 & 53.8 & \textbf{32.0} \\
\bottomrule
\end{tabular}
\end{table}

\section{Conclusion}
We presented CITRAS-FM, a tiny yet powerful TSFM that enables univariate, multivariate, and covariate-informed zero-shot forecasting with real-time CPU inference. CITRAS-FM leverages two key innovations: Shifted Attention for improved covariate alignment and CovSynth for synthesizing diverse pseudo-covariates for pretraining. CITRAS-FM achieves state-of-the-art zero-shot accuracy among sub-10M TSFMs on the fev-bench benchmark while maintaining sub-0.1-second CPU latency on covariate-informed tasks.

\bibliographystyle{IEEEtran}
\bibliography{ref}
\end{document}